\documentclass{bmvc2k}

\usepackage{xcolor}
\usepackage{multirow}
\usepackage{adjustbox}
\usepackage{pifont}
\usepackage{amsmath}
\usepackage{amssymb}
\usepackage{graphicx} 

\newcommand{\cmark}{\ding{51}}
\newcommand{\xmark}{\ding{55}}

\title{DiMoSR: Feature Modulation via Multi-Branch Dilated Convolutions for Efficient Image Super-Resolution}

\addauthor{M. Akın Yılmaz}
{mustafaakinyilmaz@ku.edu.tr}{1}
\addauthor{Ahmet Bilican}
{abilican21@ku.edu.tr}{1, 2}
\addauthor{A. Murat Tekalp}
{https://mysite.ku.edu.tr/mtekalp/}{3}

\addinstitution{Codeway AI Research\\Istanbul, TR}
\addinstitution{Koç University\\ Istanbul, TR}
\addinstitution{KUIS AI Center, Koç University\\ Istanbul, TR}

\runninghead{Yilmaz, Bilican, Tekalp}{DiMoSR}

\begin{document}

\maketitle

\begin{abstract}
Balancing reconstruction quality versus model efficiency remains a critical challenge in lightweight single image super-resolution (SISR). Despite the prevalence of attention mechanisms in recent state-of-the-art SISR approaches that primarily emphasize or suppress feature maps, alternative architectural paradigms warrant further exploration. This paper introduces DiMoSR (Dilated Modulation Super-Resolution), a novel architecture that enhances feature representation through modulation to complement attention in lightweight SISR networks. The proposed approach leverages multi-branch dilated convolutions to capture rich contextual information over a wider receptive field while maintaining computational efficiency. Experimental results demonstrate that DiMoSR outperforms state-of-the-art lightweight methods across diverse benchmark datasets, achieving superior PSNR and SSIM metrics with comparable or reduced computational complexity. Through comprehensive ablation studies, this work not only validates the effectiveness of DiMoSR but also provides critical insights into the interplay between attention mechanisms and feature modulation to guide future research in efficient network design. The code and model weights to reproduce our results are available at: \normalfont{\url{https://github.com/makinyilmaz/DiMoSR}}.
\end{abstract}

\section{Introduction}
\label{sec:intro}

Single Image Super-Resolution (SISR) aims to reconstruct high-resolution (HR) images from low-resolution (LR) inputs by recovering lost details. It has gained increased attention due to advances in streaming media and high-definition devices, creating demand for efficient super-resolution methods across resource-constrained platforms.
Deep learning has significantly advanced SISR\cite{edsrandflickr, rcan, han, dat, rdn, cansucvpr}. Leveraging the success of Transformer architectures in modeling long-range dependencies, Vision Transformer-based models have begun outperforming convolutional neural network (CNN)-based approaches\cite{swinir, swinfir, hat, drct, dat, rgt}. However, these sophisticated architectures often entail substantial computational overhead, challenging their deployment in resource-limited environments.
Numerous innovations have been proposed to reduce computational requirements\cite{ms_dilate, fdiwn, lesrcnn, fmen, elan, blueprint, srcnn, fsrcnn, lapsrn, espcn, vdsr, carn, imdn, pan, lapar, ecbsr, smsr, maffsrn, shufflemixer, rlfn, safmn, span, seemore}, focusing on decreasing model parameters and floating-point operations. In lightweight models, spatial and channel attention have been effective mechanisms to enhance feature maps by emphasizing important parts while suppressing less significant ones. However, they typically operate within limited receptive fields, creating a trade-off between contextual awareness and efficiency.
To~address these limitations, we propose DiMoSR, which leverages multi-branch dilated convolutions for feature modulation that complements traditional spatial-channel attention. Using varying dilation rates, our approach captures broader contextual information while maintaining computational efficiency, combining the strengths of both approaches in lightweight scenarios.

Our contributions can be summarized as follows.

1) We introduce DiMoSR, a lightweight SISR architecture employing dilated convolutions for effective feature modulation across varying receptive fields.

2) We demonstrate through extensive experiments that DiMoSR outperforms state-of-the-art lightweight methods across multiple datasets while maintaining comparable or reduced computational complexity.

3) We present detailed ablation studies offering insights into the interplay between feature modulation and attention mechanisms, establishing design principles for future efficient super-resolution networks.

\section{Related Work}
\label{sec:related}
\subsection{Convolutional Neural Networks for SISR}
Single Image Super-Resolution (SISR) has been revolutionized by deep learning approaches. The pioneering work SRCNN\cite{srcnn}, demonstrated the potential of convolutional neural networks (CNN) for the super-resolution task. This was followed by FSRCNN\cite{fsrcnn}, optimized for improved efficiency. VDSR\cite{vdsr} and EDSR\cite{edsrandflickr} established the benefits of increased network depth, while ESPCN\cite{espcn} introduced the efficient sub-pixel convolution operation for upsampling, now a standard component in modern SR methods.
As the field progressed, architectures became more sophisticated, with networks such as RDN\cite{rdn} leveraging dense connections, and RCAN\cite{rcan} incorporating channel attention mechanisms. However, these improvements came at the cost of increased computational complexity, limiting deployment of these models in resource-constrained environments.

\begin{figure}[t]
\centering
	\includegraphics[scale=0.17]{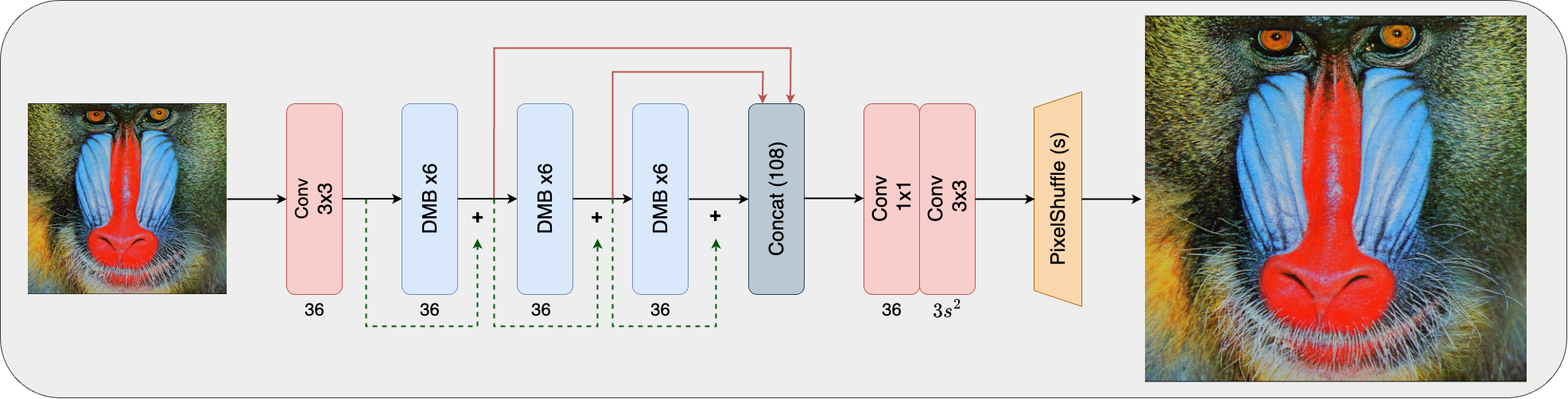} \vspace{-2pt} \\
	(a) \vspace{5pt}\\
	\includegraphics[scale=0.19]{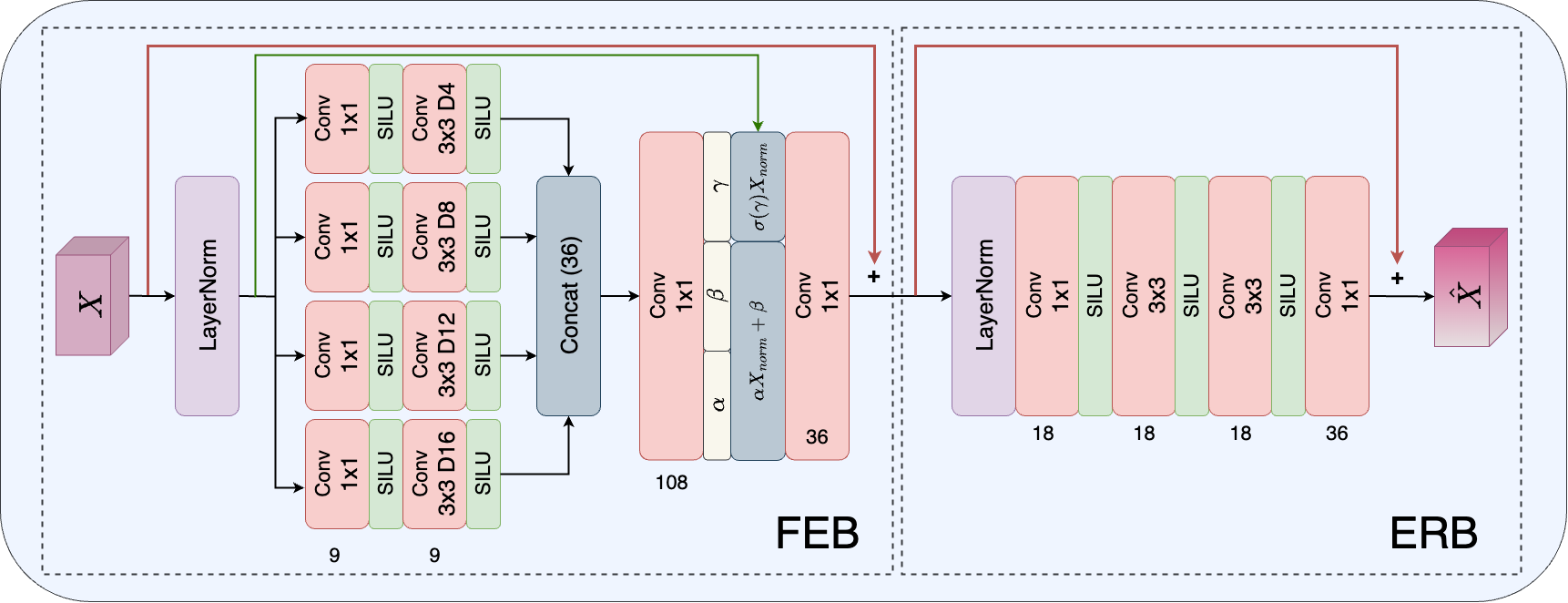} \vspace{-2pt} \\
	(b)
\caption{Our proposed DiMoSR framework: a) Architecture consisting of Dilated Modulation Blocks (DMBs), (b) Inside of a single DMB. It consists of a feature enhancement block (FEB) followed by an efficient residual block (ERB).}
\label{fig:our_model}
\end{figure}

\subsection{Lightweight Super-Resolution Networks}
In response to the computational demands of deeper and heavier networks, research has focused on developing lightweight architectures that balance reconstruction quality with efficiency. CARN\cite{carn} proposed a cascading architecture with parameter sharing to reduce model size. IMDN\cite{imdn} introduced an information multi-distillation mechanism that progressively extracts hierarchical features.
PAN\cite{pan} presented a pixel attention network that applies attention at the pixel level, offering computational advantages. ECBSR\cite{ecbsr} proposed edge-oriented convolution blocks designed for mobile devices. SMSR\cite{smsr} explored sparsity by learning location-specific convolution kernels that adaptively focus on informative regions.
Recent approaches include RLFN\cite{rlfn}, which combined local feature extraction with residual learning, LESRCNN\cite{lesrcnn}, which enhanced basic CNN architectures, FDIW\cite{fdiwn}, which introduced feature distillation interaction weighting, and FMEN\cite{fmen}, which focused on memory efficiency alongside computational performance.
\subsection{Attention Mechanisms in Super-Resolution}
Attention mechanisms have significantly influenced super-resolution model design by enabling networks to focus on informative features. While initially focused on the channel dimension in works such as RCAN\cite{rcan}, attention has evolved to encompass both spatial and channel dimensions. HAN \cite{han} introduced holistic attention that combines channel and spatial attention.
For lightweight scenarios, attention has been adapted to minimize computational overhead. MAFFSRN\cite{maffsrn} proposed multi-attention blocks with reduced complexity. SAFMN\cite{safmn} introduced spatially-adaptive feature modulation with fewer parameters. SPAN\cite{span} presented a parameter-free attention network that achieves efficiency through algorithmic innovations.

\begin{figure}[t]
  \centering
  \begin{tabular}{c@{\hspace{0em}}c}
    
    \begin{minipage}[t]{0.4\textwidth}
      \vspace*{0pt}
      \includegraphics[width=0.85\linewidth]{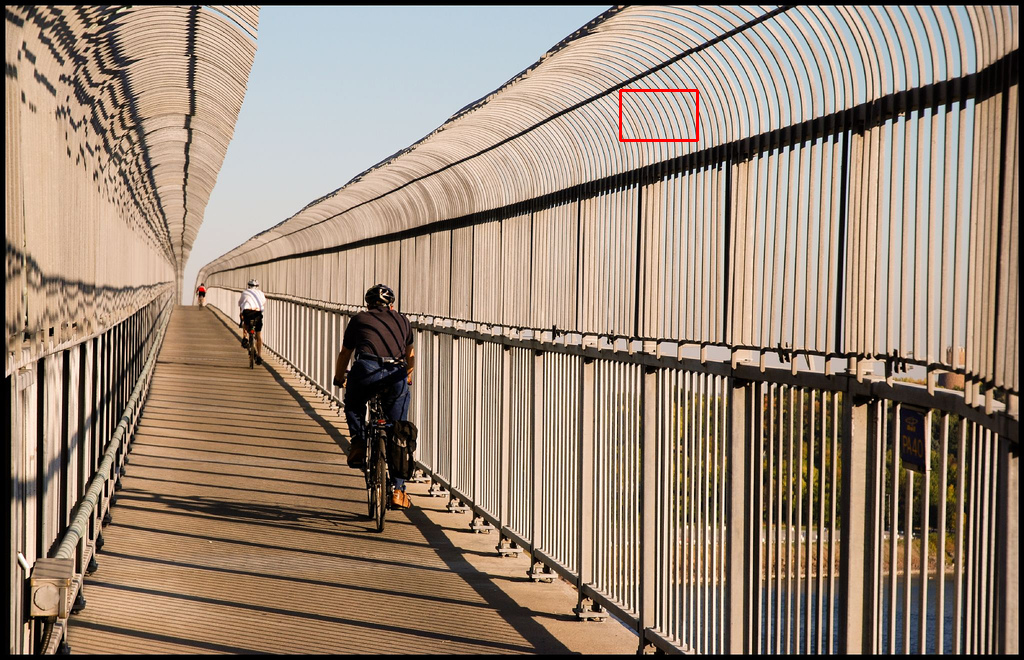}
      \\
      \centering
      \scriptsize img024 from Urban100
    \end{minipage}
    &
    \begin{minipage}[t]{0.55\textwidth}
      \vspace*{0pt}
      \centering
      \begin{tabular}{ccc}
        \includegraphics[width=0.245\linewidth]{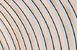} &
        \includegraphics[width=0.245\linewidth]{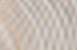} &
        \includegraphics[width=0.245\linewidth]{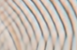} \\
        \scriptsize HR patch & \scriptsize Bicubic & \scriptsize SAFMN\cite{safmn} \\
        \includegraphics[width=0.245\linewidth]{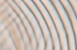} &
        \includegraphics[width=0.245\linewidth]{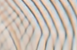} &
        \includegraphics[width=0.245\linewidth]{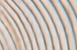} \\
        \scriptsize Shufflemixer\cite{shufflemixer} & \scriptsize SPAN\cite{span} & \scriptsize DiMoSR
      \end{tabular}
    \end{minipage}
    \\[1.5em]
    
    \begin{minipage}[t]{0.4\textwidth}
      \vspace*{0pt}
      \includegraphics[width=0.85\linewidth]{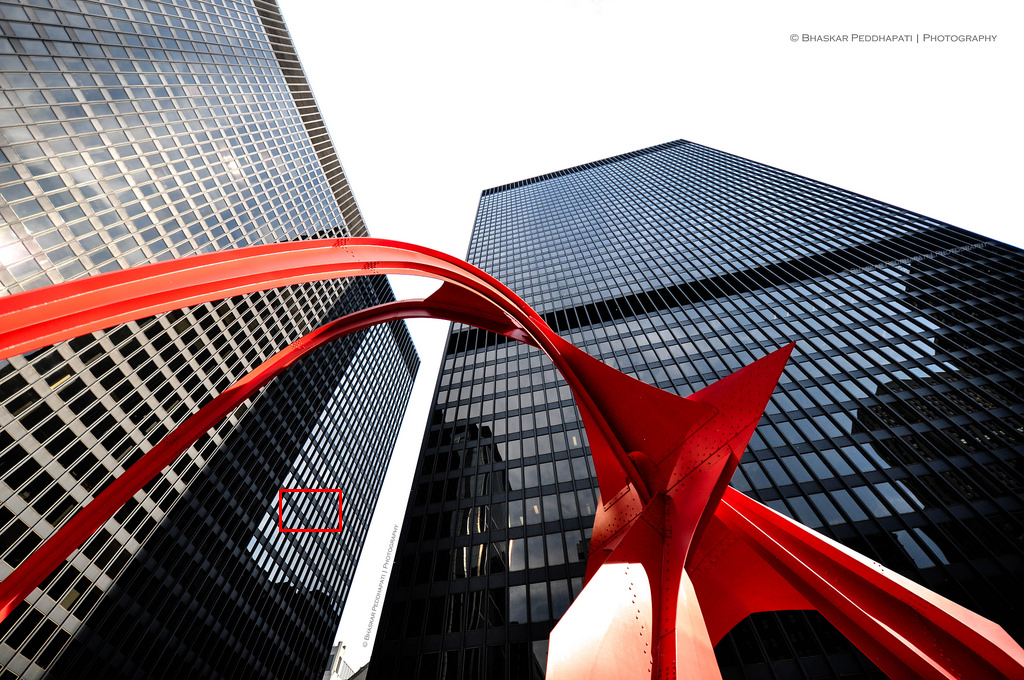}
      \\
      \centering
      \scriptsize img062 from Urban100
    \end{minipage}
    &
    \begin{minipage}[t]{0.55\textwidth}
      \vspace*{0pt}
      \centering
      \begin{tabular}{ccc}
        \includegraphics[width=0.245\linewidth]{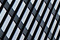} &
        \includegraphics[width=0.245\linewidth]{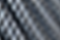} &
        \includegraphics[width=0.245\linewidth]{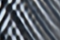} \\
        \scriptsize HR patch & \scriptsize Bicubic & \scriptsize SAFMN\cite{safmn} \\
        \includegraphics[width=0.245\linewidth]{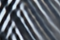} &
        \includegraphics[width=0.245\linewidth]{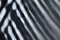} &
        \includegraphics[width=0.245\linewidth]{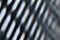} \\
        \scriptsize Shufflemixer\cite{shufflemixer} & \scriptsize SPAN\cite{span} & \scriptsize DiMoSR
      \end{tabular}
    \end{minipage}
  \end{tabular}
  \caption{Visual comparison on two Urban100 images. Left: full-size images with marked zoom-in regions. Right: HR ground truth and SR results.}
  \label{fig:urban100-comparison}
\end{figure}

\subsection{Feature Enhancement Strategies}
Recent advances have explored alternative approaches to feature enhancement beyond conventional attention mechanisms. BluePrint\cite{blueprint} introduced a separable residual network architecture that reorganizes feature channels to improve information propagation while reducing redundancy.
ShuffleMixer\cite{shufflemixer} proposed an innovative feature mixing strategy combining channel shuffling with targeted enhancement, creating diverse feature representations that capture complementary information. SAFMN\cite{safmn} introduced spatially-adaptive feature manipulation techniques that dynamically adjust representations based on local content characteristics.
While feature-wise linear modulation techniques\cite{film} have shown effectiveness in other vision tasks, their adaptation to super-resolution has been limited. Our work addresses this gap by introducing a dilated modulation framework that leverages multi-scale contextual information without significant computational overhead, providing a complementary approach particularly well-suited for resource-constrained environments.

\section{Method}
\label{sec:method}

We first provide an overview of the proposed Dilated Modulation Super-Resolution (DiMoSR) architecture for lightweight SISR, followed by detailed descriptions of its key components: the Dilated Modulation Block (DMB), the Feature Enhancement Block (FEB), and the Efficient Residual Block (ERB). 

\subsection{Overview of Network Architecture}
\label{subsec: arch}
The overall architecture of our proposed DiMoSR network is depicted in Figure\ref{fig:our_model}. Similar to most SISR networks, DiMoSR consists of three main components: (1) a shallow feature extraction module, (2) a deep feature extraction module with multiple DMBs, and (3) an image reconstruction module. Given a low-resolution (LR) input image $I_{LR} \in \mathbb{R}^{H \times W \times 3}$, we first apply a $3\times3$ convolutional layer to extract shallow features $\mathbb{R}^{H \times W \times C}$ where $C=36$ is the~number of channels in feature maps. The extracted shallow features are then fed into a deep feature extraction module consisting of a sequence of DMBs with residual connections. We organize the DMBs in a specific structure where a group of DMBs is followed by a residual connection. Specifically, after each sequence of 6 DMBs, we add the input features to the output features through a skip connection. This process is repeated 3 times, with each block's output becoming the input to the next block. After processing through all the DMB groups, the features are concatenated and passed through a $1\times1$ convolution for feature fusion. Finally, we reconstruct the super-resolved image $I_{SR} \in \mathbb{R}^{sH \times sW \times 3}$ using a sequence of $1\times1$ convolution, $3\times3$ convolution, and a pixel shuffle operation\cite{espcn} where $s$ is the upscaling factor.

\subsection{Dilated Modulation Block (DMB)}
The core of our DiMoSR architecture is the DMB, which consists of two components: the~Feature Enhancement Block (FEB) and the Efficient Residual Block (ERB). These components create a powerful mechanism for feature extraction, modulation, and attention that is both effective and computationally efficient.

\subsubsection{Feature Enhancement Block (FEB)}
The FEB, shown on the left side of Figure\ref{fig:our_model} (b), consists of multiple parallel branches with convolutional layers followed by SiLU activation functions. After a LayerNorm operation on the input features, the FEB processes them through four parallel branches, each containing a $1\times1$ convolution followed by SiLU\cite{silu} activation and a $3\times3$ dilated convolution with different dilation rates (D4, D8, D12, D16).
This multi-branch structure allows the network to capture information at different receptive fields. The outputs from all branches are concatenated into $F_{concat}$ and processed through $1\times1$  convolutions to determine the modulation coefficients $\alpha$ and $\beta$, and the attention weight $\sigma(\gamma)$ where $\sigma$ represents the sigmoid activation of $\gamma$.
The process works as follows:
\begin{equation}
\alpha, \beta, \gamma = \text{Conv}{1\times1}(F_{concat}).\text{chunk}(3)
\end{equation}
These parameters are then applied to the layer-normalized input $X_norm$ to produce modulation and attention applied outputs: 
\begin{equation}
\begin{aligned}
\text{out}1 &= \alpha X_{norm} + \beta \\
\text{out}2 &= \sigma(\gamma) X_{norm}
\end{aligned}
\end{equation}
Finally, these two outputs are concatenated and fused through another $1\times1$  convolution. A residual connection is added to complete the Feature Enhancement Block:
\begin{equation}
X_{FEB} = X + Conv_{1\times1}(Concat([out1, out2]))
\end{equation}
This combined approach allows the network to both modulate feature strengths and focus attention on the most relevant features while preserving input information through the residual connection.

\begin{table}[t]
\centering
\caption{\textbf{Benchmarking of Efficient Super-Resolution Networks
} All PSNR and SSIM metrics are calculated using only the Y-channel. \#FLOPs are measured for high-resolution output images of $1280 \times 720$ pixels. {\color{red}Red} and {\color{blue}Blue} colors indicate the best and second-best performances, respectively.}
\begin{adjustbox}{width=0.95\textwidth}
\begin{tabular}{|l|c|c|c|c|c|c|c|}
\hline
Methods       & Scale                 & \#Params {[}K{]} & \#FLOPs {[}G{]} & Set5         & Set14        & B100         & Urban100       \\ \hline
Bicubic       & \multirow{19}{*}{$\times 2$} & -                & -               & 33.66/0.9299 & 30.24/0.8688 & 29.56/0.8431 & 26.88/0.8403   \\
SRCNN\cite{srcnn}         &                       & 57               & 53              & 36.66/0.9542 & 32.42/0.9063 & 31.36/0.8879 & 29.50/0.8946   \\
FSRCNN\cite{fsrcnn}        &                       & 12               & 6               & 37.00/0.9558 & 32.63/0.9088 & 31.53/0.8920 & 29.88/0.9020   \\
ESPCN\cite{espcn}         &                       & 21               & 5               & 36.83/0.9564 & 32.40/0.9096 & 31.29/0.8917 & 29.48/0.8975   \\
VDSR\cite{vdsr}          &                       & 665              & 613             & 37.53/0.9587 & 33.03/0.9124 & 31.90/0.8960 & 30.76/0.9140   \\
LapSRN\cite{lapsrn}        &                       & 813              & 30              & 37.52/0.9590 & 33.08/0.9130 & 31.80/0.8950 & 30.41/0.9100   \\
CARN\cite{carn}          &                       & 1592             & 223             & 37.76/0.9590 & 33.52/0.9166 & 32.09/0.8978 & 31.92/0.9256   \\
EDSR-baseline\cite{edsrandflickr} &                       & 1370             & 316             & 37.99/0.9604 & 33.57/0.9175 & 32.16/0.8994 & 31.98/0.9272   \\
IMDN\cite{imdn}          &                       & 694              & 159             & 38.00/0.9605 & 33.63/0.9177 & 32.19/0.8996 & 32.17/0.9283   \\
PAN\cite{pan}           &                       & 261              & 71              & 38.00/0.9605 & 33.59/0.9181 & 32.18/0.8997 & 32.01/0.9273   \\
LAPAR-A\cite{lapar}       &                       & 548              & 171             & 38.01/0.9605 & 33.62/0.9183 & 32.19/0.8999 & 32.10/0.9283   \\
MAFFSRN\cite{maffsrn}       &                       & 402              & 86              & 37.97/0.9603 & 33.49/0.9170 & 32.14/0.8994 & 31.96/0.9268   \\
SMSR\cite{smsr}          &                       & 985              & 132             & 38.00/0.9601 & 33.64/0.9179 & 32.17/0.8990 & 32.19/0.9284   \\
ECBSR\cite{ecbsr}         &                       & 596              & 137             & 37.90/{\color{red}0.9615} & 33.34/0.9178 & 32.10/{\color{red}0.9018} & 31.71/0.9250   \\
RLFN\cite{rlfn}          &                       & 527              & 116             & {\color{blue}38.07}/0.9607 & \color{blue}33.72/0.9187 & {\color{blue}32.22}/0.9000 & \color{red}32.33/0.9299   \\
ShuffleMixer\cite{shufflemixer}  &                       & 394              & 91              & 38.01/0.9606 & 33.63/0.9180 & 32.17/0.8995 & 31.89/0.9257   \\
SAFMN\cite{safmn}         &                       & 228              & 52              & 38.00/0.9605 & 33.54/0.9177 & 32.16/0.8995 & 31.84/0.9256   \\
SPAN\cite{span}          &                       & 481              &   94              & {\color{red}38.08}/{\color{blue}0.9608} & 33.71/0.9183 & {\color{blue}32.22}/0.9002 & 32.24/0.9294 \\
\textbf{DiMoSR-S} &                       & \textbf{239}              & \textbf{54}              & 38.02/0.9606 & 33.67/0.9185 & {\color{blue}32.22}/0.9002 & 32.16/0.9283   \\
\textbf{DiMoSR (Ours)} &                       & \textbf{338}              & \textbf{76}              & 38.06/0.9607 & \color{red}33.74/0.9194 & {\color{red}32.24}/{\color{blue}0.9006} & \color{blue}32.30/0.9295   \\ \hline
Bicubic       & \multirow{19}{*}{$\times 4$} & -                & -               & 28.42/0.8104 & 26.00/0.7027 & 25.96/0.6675 & 23.14/0.6577   \\
SRCNN\cite{srcnn}         &                       & 57               & 53              & 30.48/0.8628 & 27.49/0.7503 & 26.90/0.7101 & 24.52/0.7221   \\
FSRCNN\cite{fsrcnn}        &                       & 12               & 5               & 30.71/0.8657 & 27.59/0.7535 & 26.98/0.7150 & 24.62/0.7280   \\
ESPCN\cite{espcn}         &                       & 25               & 1               & 30.52/0.8697 & 27.42/0.7606 & 26.87/0.7216 & 24.39/0.7241   \\
VDSR\cite{vdsr}          &                       & 665              & 613             & 31.35/0.8838 & 28.01/0.7674 & 27.29/0.7251 & 25.18/0.7524   \\
LapSRN\cite{lapsrn}        &                       & 813              & 149             & 31.54/0.8850 & 28.19/0.7720 & 27.32/0.7280 & 25.21/0.7560   \\
CARN\cite{carn}          &                       & 1592             & 91              & 32.13/0.8937 & 28.60/0.7806 & 27.58/0.7349 & 26.07/0.7837   \\
EDSR-baseline\cite{edsrandflickr} &                       & 1518             & 114             & 32.09/0.8938 & 28.58/0.7813 & 27.57/0.7357 & 26.04/0.7849   \\
IMDN\cite{imdn}          &                       & 715              & 41              & 32.21/0.8948 & 28.58/0.7811 & 27.56/0.7353 & 26.04/0.7838   \\
PAN\cite{pan}           &                       & 272              & 28              & 32.13/0.8948 & 28.61/0.7822 & 27.59/0.7363 & 26.11/0.7854   \\
LAPAR-A\cite{lapar}       &                       & 659              & 94              & 32.15/0.8944 & 28.61/0.7818 & 27.61/0.7366 & 26.14/0.7871   \\
MAFFSRN\cite{maffsrn}       &                       & 441              & 24              & 32.18/0.8948 & 28.58/0.7812 & 27.57/0.7361 & 26.04/0.7848   \\
SMSR\cite{smsr}          &                       & 1006             & 42              & 32.12/0.8932 & 28.55/0.7808 & 27.55/0.7351 & 26.11/0.7868   \\
ECBSR\cite{ecbsr}         &                       & 603              & 35              & 31.92/0.8946 & 28.34/0.7817 & 27.48/{\color{red}0.7393} & 25.81/0.7773   \\
RLFN\cite{rlfn}          &                       & 543              & 30              & {\color{blue}32.24}/0.8952 & 28.62/0.7813 & 27.60/0.7364 & 26.17/0.7877   \\
ShuffleMixer\cite{shufflemixer}  &                       & 411              & 28              & 32.21/{\color{blue}0.8953} & {\color{blue}28.66}/0.7827 & 27.61/0.7366 & 26.08/0.7835   \\
SAFMN\cite{safmn}         &                       & 240              & 14              & 32.18/0.8948 & 28.60/0.7813 & 27.58/0.7359 & 25.97/0.7809   \\
SPAN\cite{span}          &                       & 498              &   24              & 32.20/{\color{blue}0.8953} & \color{blue}28.66/0.7834 & {\color{blue}27.62}/0.7374 & \color{blue}26.18/0.7879   \\
\textbf{DiMoSR-S} &                       & \textbf{250}              & \textbf{14}              & {\color{blue}32.24}/0.8952 & {\color{blue}28.66}/0.7827 & {\color{blue}27.62}/0.7374 & 26.12/0.7847   \\
\textbf{DiMoSR} &                       & \textbf{349}              & \textbf{20}              & \color{red}32.31/0.8962 & \color{red}28.74/0.7844 & {\color{red}27.65}/{\color{blue}0.7384} & \color{red}26.25/0.7889   \\ \hline

\end{tabular}
\end{adjustbox}
\label{table:results}
\end{table}

\subsubsection{Efficient Residual Block (ERB)}
The right side of Figure\ref{fig:our_model} (b) shows our Efficient Residual Block (ERB), which takes the~output from the FEB as its input. The ERB follows a residual block structure for effective feature processing. After a LayerNorm operation on the input features, the ERB first maps them to a lower channel dimension using a $1\times1$ convolution for improved efficiency. These reduced features are then processed through $3\times3$ convolution layers, and finally, another $1\times1$ convolution is applied to increase the channel count back to match the input feature dimensions. Similar to the FEB, SiLU activation functions are applied between each convolution layer.

The ERB process can be formulated as:
\begin{equation}
\hat{X} = X_{FEB} + \text{ERB}(\text{LayerNorm}(X_{FEB})),
\end{equation}
where $X_{FEB}$ represents the features from the Feature Enhancement Block and $\text{ERB}(\cdot)$ represents the operations performed by the Efficient Residual Block. The final residual connection helps preserve information flow and stabilize training.

The complementary ERB pathway provides a computationally efficient feature processing stream that works alongside the more complex attention and modulation mechanisms of the FEB. While the FEB focuses on applying attention and modulation operations to enhance features, the ERB handles the direct feature processing through its efficient design.

\section{Experiments}
\label{sec:experiments}

In this section, we evaluate our proposed DiMoSR method through comprehensive experiments. We first outline the experimental setup, followed by comparisons with state-of-the-art methods and ablation studies where we analyze individual components of our approach.
\subsection{Experimental Setup}
\label{subsec}
\paragraph{Datasets.}
For training, we use DF2K dataset which is a combination of DIV2K\cite{div2k} and Flickr2K\cite{edsrandflickr} datasets, comprising a total of 3450 high-quality images with diverse content. Low-resolution (LR) inputs are generated by applying bicubic downscaling to the high-resolution (HR) reference images. We evaluate our model on four benchmark datasets: Set5\cite{set5}, Set14\cite{set14}, B100\cite{b100}, and Urban100\cite{urban100}, representing a range of image types from natural scenes to urban environments.

\begin{table}[t]
\centering
\caption{Effect of attention mechanism and feature modulation components on $\times4$ upscaling. Results are presented as PSNR/SSIM across four benchmark datasets.}
\begin{adjustbox}{width=0.8\textwidth}
\begin{tabular}{cccccc}
\hline
Attention & \begin{tabular}[c]{@{}c@{}}Feature\\ Modulation\end{tabular} & Set5 & Set14 & B100 & Urban100 \\ \hline
\xmark & \xmark & 31.94/0.8920 & 28.50/0.7795 & 27.51/0.7336 & 25.80/0.7743 \\ \hline
\cmark & \xmark & 32.28/0.8958 & 28.69/0.7834 & 27.63/0.7376 & 26.18/0.7867 \\ \hline
\xmark & \cmark & 32.26/0.8959 & 28.69/0.7836 & 27.63/0.7376 & 26.14/0.7858 \\ \hline
\cmark & \cmark & \textbf{32.31/0.8962} & \textbf{28.74/0.7844} & \textbf{27.65/0.7384} & \textbf{26.25/0.7889} \\ \hline
\end{tabular}
\end{adjustbox}
\label{table:ablation}
\end{table}

\paragraph{Evaluation Metrics.}
Following standard practice, we measure reconstruction quality using Peak Signal-to-Noise Ratio (PSNR) and Structural Similarity Index (SSIM)\cite{ssim}. All metrics are calculated on the Y channel in YCbCr color space.

\paragraph{Training Details.}
For training both our DiMoSR ($36$ channels and $18$ DMBs) and its smaller sibling DiMoSR-S ($32$ channels and $16$ DMBs), we applied data augmentation including random horizontal flips and 90°, 180°, and 270° rotations. We randomly cropped patches of size $128 \times 128$ pixels from LR images, corresponding to $512 \times 512$ patches in HR images for $\times 4$ upscaling. We optimized our models using Adam\cite{adam} optimizer with $\beta_1 = 0.9$ and $\beta_2 = 0.99$ for 500,000 iterations with a batch size of 24 pairs. The learning rate starts at $1 \times 10^{-3}$ and decreases to $1 \times 10^{-5}$ using Cosine Annealing\cite{cosineanneal}. Following\cite{timofte_fourier, shufflemixer, safmn}, our loss function combines mean absolute error (MAE) and FFT-based frequency loss:
\begin{equation}
\mathcal{L}_{total} = \mathcal{L}_{MAE} + \lambda \mathcal{L}_{freq}
\end{equation}
\noindent where $\mathcal{L}_{MAE} = |I_{SR} - I_{HR}|$ is the pixel-wise MAE, $\mathcal{L}_{freq} = |\mathcal{F}(I_{SR}) - \mathcal{F}(I_{HR})|$ is the~frequency domain loss computed using Fast Fourier Transform (FFT), and $\lambda = 0.05$ balances the two terms.

\subsection{Comparison with State-of-the-Art Methods} We compare our DiMoSR method with various efficient super-resolution approaches across standard benchmark datasets for scale factors $\times 2$, $\times 4$ and present quantitative results in Table\ref{table:results}. For $\times 2$ upscaling, DiMoSR outperforms many models with significantly higher parameter counts and computational costs, providing better PSNR values than RLFN, ShuffleMixer, and SPAN on several benchmarks.

Our performance is particularly notable for $\times 4$ upscaling, where DiMoSR achieves state-of-the-art performance on all four benchmark datasets in terms of PSNR and on three of the four benchmark datasets in terms of SSIM. These results are achieved with significant efficiency gains compared to recent efficient SR methods that have more parameter count and FLOPs.

In particular, our lightweight variant DiMoSR-S for $\times$2 upscaling substantially outperforms SAFMN across all benchmark datasets despite having nearly identical computational requirements. For $\times$4 upscaling, DiMoSR-S achieves remarkable results on Set5 (32.24 dB), matching RLFN for the second-best performance while utilizing 54\% fewer parameters and 53\% fewer FLOPs. On Set14, DiMoSR-S (28.66 dB) achieves equivalent performance to ShuffleMixer and SPAN, securing joint second place despite requiring significantly reduced computational complexity. Compared to these competing architectures, DiMoSR-S demonstrates substantial efficiency gains of 39-50\% in parameter count and 41.7-50\% in floating-point operations while maintaining comparable or superior quantitative results across multiple benchmark datasets, including B100 (27.62 dB). These empirical findings conclusively demonstrate that our approach successfully addresses the fundamental challenge of achieving superior reconstruction quality while simultaneously reducing computational requirements.

\begin{figure}[t]
  \centering
  \begin{tabular}{c@{\hspace{0em}}c}
    
    \begin{minipage}[t]{0.4\textwidth}
      \vspace*{0pt}
      \includegraphics[width=0.85\linewidth]{images/gt062_with_rect.png}
      \\
      \centering
      \scriptsize img062 from Urban100
    \end{minipage}
    &
    \begin{minipage}[t]{0.55\textwidth}
      \vspace*{0pt}
      \centering
      \begin{tabular}{ccc}
        \includegraphics[width=0.245\linewidth]{images/gt062.png} &
        \includegraphics[width=0.245\linewidth]{images/bicubic062.png} &
        \includegraphics[width=0.245\linewidth]{images/safmn062.png} \\
        \scriptsize HR patch & \scriptsize Bicubic & \scriptsize SAFMN\cite{safmn} \\
        \includegraphics[width=0.245\linewidth]{images/shufflemixer062.png} &
        \includegraphics[width=0.245\linewidth]{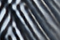} &
        \includegraphics[width=0.245\linewidth]{images/ours062.png} \\
        \scriptsize Shufflemixer\cite{shufflemixer} & \scriptsize Ours (wo FFT) & \scriptsize Ours
      \end{tabular}
    \end{minipage}
\end{tabular}
  \caption{Visual comparison on img062 from Urban100 ($\times 4$ upscaling). Note that SAFMN\cite{safmn} and ShuffleMixer\cite{shufflemixer} were trained with FFT loss, yet our model without FFT loss achieves comparable results. Our complete model with FFT Loss demonstrates superior edge preservation and detail reconstruction, particularly in structural patterns.}
  \label{fig:lossabl}
\end{figure}

\subsection{Ablation Study}
\subsubsection{Feature Modulation and Attention}
To validate both the attention mechanism and feature modulation components of our DiMoSR model we conduct comprehensive ablation studies analyzing the individual and combined contributions of the attention mechanism and feature modulation components in Table\ref{table:ablation}.
When either component is disabled independently, we observe performance decreases across all benchmark datasets for $\times4$ upscaling. The most significant impact appears on the Urban100 dataset, where disabling attention causes a 0.10dB decrease compared to our full model. Disabling feature modulation similarly reduces performance across all datasets.
When both components are enabled, we achieve the best performance across all datasets, demonstrating that attention and feature modulation mechanisms contribute significantly to DiMoSR's overall performance. Their combined effect is particularly beneficial for images with complex structures, as evidenced on the Urban100 dataset.

\subsubsection{FFT Loss}

To evaluate the effectiveness of the frequency domain loss function, we conduct an ablation study on the impact of FFT loss in our DiMoSR model. Table\ref{table:loss} presents a comparison between training with standard MAE only versus combining MAE with FFT loss using a weighting factor of 0.05. We observe notable performance improvements for $\times 4$ upscaling across all benchmark datasets when introducing the FFT loss component along with the standard MAE. 

Figure\ref{fig:lossabl} provides visual evidence of the FFT loss impact on reconstruction quality. While our base model without FFT loss already performs comparably to other state-of-the-art models that utilize FFT loss (SAFMN and ShuffleMixer), integrating FFT loss into our DiMoSR model produces substantially improved visual results. The zoomed-in regions clearly demonstrate that adding FFT loss helps recover more accurate high-frequency details and sharper edges, particularly in structures with complex patterns, as found in urban scenes. These results demonstrate that the frequency domain loss complements the spatial domain supervision provided by the MAE, resulting in better visual quality and quantitative metrics across diverse image content.

\begin{table}[t]
\centering
\caption{Effect of FFT loss on $\times4$ upscaling. Results are presented as PSNR/SSIM across four benchmark datasets.}
\begin{adjustbox}{width=0.8\textwidth}
\begin{tabular}{cccccc}
\hline
MAE & \begin{tabular}[c]{@{}c@{}}FFT\\Loss\end{tabular} & Set5 & Set14 & B100 & Urban100 \\ \hline
\cmark & \xmark & 32.26/0.8957 & 28.65/0.7829 & 27.62/0.7375 & 26.15/0.7884 \\ \hline
\cmark & \cmark & \textbf{32.31/0.8962} & \textbf{28.74/0.7844} & \textbf{27.65/0.7384} & \textbf{26.25/0.7889} \\ \hline
\end{tabular}
\end{adjustbox}
\label{table:loss}
\end{table}

\section{Conclusion}
\label{sec:conclusion}
This paper introduces DiMoSR, a novel lightweight super-resolution network that leverages multi-branch dilated convolutions and feature modulation to achieve efficient and high-quality image super-resolution. Extensive experiments on benchmark datasets demonstrate that DiMoSR consistently outperforms state-of-the-art lightweight SR methods, particularly for the challenging $\times4$ upscaling task. Even our smaller variant, DiMoSR-S, delivers competitive performance while requiring significantly fewer computational resources than comparable methods. Both models achieve superior PSNR and SSIM metrics across multiple datasets compared to alternatives with higher computational demands. Our ablation studies confirm that both attention and feature modulation components contribute significantly to the overall performance, with their combined effect showing particular benefits for complex image structures.
The success of DiMoSR highlights the potential of alternative feature enhancement approaches beyond traditional attention mechanisms for lightweight super-resolution networks.

\clearpage
\bibliography{egbib}
\end{document}